\pdfoutput=1

\documentclass[11pt,a4paper]{article}
\usepackage{authblk}
\usepackage[hyperref]{acl2021}
\usepackage{times}
\usepackage{latexsym}

\usepackage{microtype}

\usepackage{textcomp}
\usepackage{multirow}
\usepackage{graphicx}
\usepackage{xcolor}
\usepackage{array}
\usepackage{booktabs}
\definecolor{green}{RGB}{153,255,153}
\definecolor{hred}{RGB}{255,153,153}
\usepackage{soul}
\DeclareRobustCommand{\hlgreen}[1]{\sethlcolor{green}\hl{#1}}
\DeclareRobustCommand{\hlred}[1]{\sethlcolor{hred}\hl{#1}}
\newcommand{\PreserveBackslash}[1]{\let\temp=\\#1\let\\=\temp}
\newcolumntype{C}[1]{>{\PreserveBackslash\centering}p{#1}}
\usepackage [english]{babel}
\usepackage [autostyle, english = american]{csquotes}
\MakeOuterQuote{"}

\aclfinalcopy 

\title{\textsc{HateCheck}: Functional Tests for Hate Speech Detection Models}

\author[1,2]{\textbf{Paul Röttger}}
\author[2]{\textbf{Bertram Vidgen}}
\author[3]{\textbf{Dong Nguyen}}
\author[4]{\textbf{Zeerak Waseem}}
\author[1,2]{\\\textbf{Helen Margetts}}
\author[1]{\textbf{Janet B. Pierrehumbert}}
\affil[1]{University of Oxford}
\affil[2]{The Alan Turing Institute}
\affil[3]{Utrecht University}
\affil[4]{University of Sheffield}

\begin{document}
\maketitle
\begin{abstract}
Detecting online hate is a difficult task that even state-of-the-art models struggle with.
Typically, hate speech detection models are evaluated by measuring their performance on held-out test data using metrics such as accuracy and F1 score.
However, this approach makes it difficult to identify specific model weak points.
It also risks overestimating generalisable model performance due to increasingly well-evidenced systematic gaps and biases in hate speech datasets.
To enable more targeted diagnostic insights, we introduce \textsc{HateCheck}, a suite of functional tests for hate speech detection models.
We specify 29 model functionalities motivated by a review of previous research and a series of interviews with civil society stakeholders.
We craft test cases for each functionality and validate their quality through a structured annotation process.
To illustrate \textsc{HateCheck}'s utility, we test near-state-of-the-art transformer models as well as two popular commercial models, revealing critical model weaknesses.
\end{abstract}

\section{Introduction} \label{sec: intro}
Hate speech detection models play an important role in online content moderation and enable scientific analyses of online hate more generally.
This has motivated much research in NLP and the social sciences.
However, even state-of-the-art models exhibit substantial weaknesses
\citep[see][for reviews]{schmidt2017survey,fortuna2018survey,vidgen2019challenges, mishra2020tackling}.

So far, hate speech detection models have primarily been evaluated by measuring held-out performance on a small set of widely-used hate speech datasets \citep[particularly][]{waseem2016hateful, davidson2017automated,founta2018large}, but recent work has highlighted the limitations of this evaluation paradigm.
Aggregate performance metrics offer limited insight into specific model weaknesses \citep{wu2019errudite}.
Further, if there are systematic gaps and biases in training data, models may perform deceptively well on corresponding held-out test sets by learning simple decision rules rather than encoding a more generalisable understanding of the task \citep[e.g.][]{niven2019probing,geva2019modeling,shah2020predictive}.
The latter issue is particularly relevant to hate speech detection since current hate speech datasets vary in data source, sampling strategy and annotation process \citep{vidgen2020directions,poletto2020resources}, and are known to exhibit annotator biases \citep{waseem2016you,waseem2018bridging,sap2019risk} as well as topic and author biases \citep{wiegand2019detection, nejadgholi2020cross}.
Correspondingly, models trained on such datasets have been shown to be overly sensitive to lexical features such as group identifiers \citep{park2018reducing, dixon2018measuring, kennedy2020contextualizing}, and to generalise poorly to other datasets \citep{nejadgholi2020cross,samory2020unsex}.
Therefore, held-out performance on current hate speech datasets is an incomplete and potentially misleading measure of model quality.

To enable more targeted diagnostic insights, we introduce \textsc{HateCheck}, a suite of functional tests for hate speech detection models.
Functional testing, also known as black-box testing, is a testing framework from software engineering that assesses different functionalities of a given model by validating its output on sets of targeted test cases \citep{beizer1995black}.
\citet{ribeiro2020beyond} show how such a framework can be used for structured model evaluation across diverse NLP tasks.

\textsc{HateCheck} covers 29 model functionalities, the selection of which we motivate through a series of interviews with civil society stakeholders and a review of hate speech research.
Each functionality is tested by a separate functional test.
We create 18 functional tests corresponding to distinct expressions of hate.
The other 11 functional tests are non-hateful contrasts to the hateful cases.
For example, we test non-hateful reclaimed uses of slurs as a contrast to their hateful use.
Such tests are particularly challenging to models relying on overly simplistic decision rules and thus enable more accurate evaluation of true model functionalities \citep{gardner2020evaluating}.
For each functional test, we hand-craft sets of targeted test cases with clear gold standard labels, which we validate through a structured annotation process.\footnote{All \textsc{HateCheck} test cases and annotations are available on \href{https://github.com/paul-rottger/hatecheck-data}{https://github.com/paul-rottger/hatecheck-data}.}

\textsc{HateCheck} is broadly applicable across English-language hate speech detection models.
We demonstrate its utility as a diagnostic tool by evaluating two BERT models \citep{devlin2019bert}, which have achieved near state-of-the-art performance on hate speech datasets \citep{tran2020habertor}, as well as two commercial models -- Google Jigsaw's \href{https://www.perspectiveapi.com}{Perspective} and Two Hat's \href{https://www.siftninja.com/}{SiftNinja}.\footnote{\href{https://www.perspectiveapi.com}{www.perspectiveapi.com} and \href{https://www.siftninja.com/}{www.siftninja.com}}
When tested with \textsc{HateCheck}, all models appear overly sensitive to specific keywords such as slurs.
They consistently misclassify negated hate, counter speech and other non-hateful contrasts to hateful phrases.
Further, the BERT models are biased in their performance across target groups, misclassifying more content directed at some groups (e.g. women) than at others.
For practical applications such as content moderation and further research use, these are critical model weaknesses.
We hope that by revealing such weaknesses, \textsc{HateCheck} can play a key role in the development of better hate speech detection models.

\paragraph{Definition of Hate Speech}
We draw on previous definitions of hate speech \citep{warner2012detecting,davidson2017automated} as well as recent typologies of abusive content \citep{vidgen2019challenges, banko2020unified} to define hate speech as \textit{abuse that is targeted at a protected group or at its members for being a part of that group}.
We define protected groups based on age, disability, gender identity, familial status, pregnancy, race, national or ethnic origins, religion, sex or sexual orientation, which broadly reflects international legal consensus (particularly the UK's 2010 Equality Act, the US 1964 Civil Rights Act and the EU's Charter of Fundamental Rights).
Based on these definitions, we approach hate speech detection as the binary classification of content as either hateful or non-hateful.
Other work has further differentiated between different types of hate and non-hate \citep[e.g.][]{founta2018large,salminen2018anatomy,zampieri2019predicting}, but such taxonomies can be collapsed into a binary distinction and are thus compatible with \textsc{HateCheck}.

\paragraph{Content Warning}
This article contains examples of hateful and abusive language. 
All examples are taken from \textsc{HateCheck} to illustrate its composition.
Examples are quoted verbatim, except for hateful slurs and profanity, for which the first vowel is replaced with an asterisk.

\section{\textsc{HateCheck}} \label{sec: test suite construction}
\subsection{Defining Model Functionalities} \label{subsec: func definition}

In software engineering, a program has a certain \textit{functionality} if it meets a specified input/output behaviour \citep{ISO24765}.
Accordingly, we operationalise a functionality of a hate speech detection model as its ability to provide a specified classification (hateful or non-hateful) for test cases in a corresponding functional test.
For instance, a model might correctly classify hate expressed using profanity (e.g "F*ck all black people") but misclassify non-hateful uses of profanity (e.g. "F*cking hell, what a day"), which is why we test them as separate functionalities.
Since both functionalities relate to profanity usage, we group them into a common \textit{functionality class}.

\subsection{Selecting Functionalities for Testing} \label{subsec: selecting functionalities}

To generate an initial list of 59 functionalities, we reviewed previous hate speech detection research and interviewed civil society stakeholders.

\paragraph{Review of Previous Research}
We identified different types of hate in taxonomies of abusive content \citep[e.g.][]{zampieri2019predicting, banko2020unified, kurrek2020comprehensive}.
We also identified likely model weaknesses based on error analyses \citep[e.g.][]{davidson2017automated, van2018challenges, vidgen2020detecting} as well as review articles and commentaries \citep[e.g.][]{schmidt2017survey, fortuna2018survey, vidgen2019challenges}.
For example, hate speech detection models have been shown to struggle with correctly classifying negated phrases such as "I don't hate trans people" \citep{hosseini2017deceiving, dinan2019build}.
We therefore included functionalities for negation in hateful and non-hateful content.

\paragraph{Interviews}
We interviewed 21 employees from 16 British, German and American NGOs whose work directly relates to online hate.
Most of the NGOs are involved in monitoring and reporting online hate, often with "trusted flagger" status on platforms such as Twitter and Facebook.
Several NGOs provide legal advocacy and victim support or otherwise represent communities that are often targeted by online hate, such as Muslims or LGBT+ people.
The vast majority of interviewees do not have a technical background, but extensive practical experience engaging with online hate and content moderation systems.
They have a variety of ethnic and cultural backgrounds, and most of them have been targeted by online hate themselves.

The interviews were semi-structured.
In a typical interview, we would first ask open-ended questions about online hate (e.g. "What do you think are the biggest challenges in tackling online hate?") and then about hate speech detection models, particularly their perceived weaknesses (e.g. "What sort of content have you seen moderation systems get wrong?") and potential improvements, unbounded by technical feasibility (e.g. "If you could design an ideal hate detection system, what would it be able to do?").
Using a grounded theory approach \citep{corbin1990grounded}, we identified emergent themes in the interview responses and translated them into model functionalities.
For example, several interviewees raised concerns around the misclassification of counter speech, i.e. direct responses to hateful content (e.g. I4: "people will be quoting someone, calling that person out [...] but that will get picked up by the system").\footnote{When quoting anonymised responses throughout this article, we identify each interview participant by a unique ID. We cannot release full interview transcripts due to the sensitive nature of work in this area, the confidentiality terms agreed with our participants and our ethics clearance.}
We therefore included functionalities for counter speech that quotes or references hate.

\paragraph{Selection Criteria}
From the initial list of 59 functionalities, we select those in \textsc{HateCheck} based on two practical considerations.

First, we restrict \textsc{HateCheck}'s scope to individual English language text documents.
This is due to practical constraints, and because most hate speech detection models are developed for such data \citep{poletto2020resources,vidgen2020directions}.
Thus, \textsc{HateCheck} does not test functionalities that relate to other modalities (e.g. images) or languages, or that require context (e.g. conversational or social) beyond individual documents.

Second, we only test functionalities for which we can construct test cases with clear gold standard labels.
Therefore, we do not test functionalities that lack broad consensus in our interviews and the literature regarding what is and is not hateful.
The use of humour, for instance, has been highlighted as an important challenge for hate speech research \citep{van2018challenges,qian2018leveraging,vidgen2020detecting}.
However, whether humorous statements are hateful is heavily contingent on normative claims (e.g. I5: "it's a value judgment thing"), which is why we do not test them in \textsc{HateCheck}.

\subsection{Functional Tests in \textsc{HateCheck}} \label{subsec: func in test suite}

\textsc{HateCheck} comprises 29 functional tests grouped into 11 classes.
Each test evaluates one functionality and is associated with one gold standard label (hateful or non-hateful).
Each functional test has a set of corresponding test cases.

18 functional tests for hateful content in \textsc{HateCheck} cover \textbf{distinct expressions of hate}.
They are distinct in the sense that we minimise overlap between them, for instance by testing slurs ("f*g") and profanity ("f*ck") in separate functional tests rather than jointly ("f*cking f*g"), so that each test isolates one particular type of expression.

The other 11 functional tests for non-hateful content cover \textbf{contrastive non-hate}, i.e. content which shares linguistic features with hateful expressions.
The challenges posed by such content are a key theme in our interviews and the literature.
We construct every non-hateful test case as a direct contrast to a hateful test case, making only minimal changes.
For instance, "I love immigrants" is a test case in \textbf{F19}: positive statements using a protected group identifier.
It directly contrasts the test case "I hate immigrants" in \textbf{F1}: strong negative emotions explicitly expressed about a protected group.

In the following, we give a brief overview of the different functional tests in \textsc{HateCheck}.
Table \ref{fig: functionalities} provides corresponding example test cases.
Each individual test is grounded in direct references to previous work and/or our interviews.
These references are detailed in Appendix \ref{app: functional tests}.

\subsubsection*{Distinct Expressions of Hate}

\textsc{HateCheck} tests different types of derogatory hate speech (\textbf{F1-4}) and hate expressed through threatening language (\textbf{F5/6}).
It tests hate expressed using slurs (\textbf{F7}) and profanity (\textbf{F10}).
It also tests hate expressed through pronoun reference (\textbf{F12/13}), negation (\textbf{F14)} and phrasing variants, specifically questions and opinions (\textbf{F16/17}).
Lastly, it tests hate containing spelling variations such as missing characters or leet speak (\textbf{F25-29}).

\subsubsection*{Contrastive Non-Hate}

\textsc{HateCheck} tests non-hateful contrasts for slurs, particularly slur homonyms and reclaimed slurs (\textbf{F8/9}), as well as for profanity (\textbf{F11}).
It tests non-hateful contrasts that use negation, i.e. negated hate (\textbf{F15}).
It also tests non-hateful contrasts around protected group identifiers (\textbf{F18/19}).
It tests contrasts in which hate speech is quoted or referenced to non-hateful effect, specifically \textit{counter speech}, i.e. direct responses to hate speech which seek to act against it (\textbf{F20/21}).
Lastly, it tests non-hateful contrasts which target out-of-scope entities such as objects (\textbf{F22-24}) rather than a protected group.

\subsection{Generating Test Cases} \label{subsec: generating test cases}

For each functionality in \textsc{HateCheck}, we hand-craft sets of test cases -- short English-language text documents that clearly correspond to just one gold standard label.
Within each functionality, we aim to use diverse vocabulary and syntax to reduce similarity between test cases, which \citet{zhou2020curse} suggest as a likely cause of performance instability for diagnostic datasets.

To generate test cases at scale, we use templates \citep{dixon2018measuring,garg2019counterfactual,ribeiro2020beyond}, in which we replace tokens for protected group identifiers (e.g. "I hate [IDENTITY].") and slurs (e.g. "You are just a [SLUR] to me.").
This also ensures that \textsc{HateCheck} has an equal number of cases targeted at different protected groups.

\textsc{HateCheck} covers seven protected groups:
women (gender), trans people (gender identity), gay people (sexual orientation), black people (race), disabled people (disability), Muslims (religion) and immigrants (national origin). 
For details on which slurs are covered by \textsc{HateCheck} and how they were selected, see Appendix \ref{app: slurs}.

In total, we generate 3,901 cases, 3,495 of which come from 460 templates.
The other 406 cases do not use template tokens (e.g. "Sh*t, I forgot my keys") and are thus crafted individually.
The average length of cases is 8.87 words (std. dev.~=~3.33) or 48.26 characters (std. dev.~=~16.88).
2,659 of the 3,901 cases (68.2\%) are hateful and 1,242 (31.8\%) are non-hateful.

\paragraph{Secondary Labels}
In addition to the primary label (hateful or non-hateful) we provide up to two secondary labels for all cases.
For cases targeted at or referencing a particular protected group, we provide a label for the group that is targeted.
For hateful cases, we also label whether they are targeted at a group in general or at individuals, which is a common distinction in taxonomies of abuse \citep[e.g.][]{waseem2017understanding,zampieri2019predicting}.

\subsection{Validating Test Cases} \label{subsec: validating test cases}

To validate gold standard primary labels of test cases in \textsc{HateCheck}, we recruited and trained ten annotators.\footnote{For information on annotator training, their background and demographics, see the data statement in Appendix \ref{app: data statement}.} 
In addition to the binary annotation task, we also gave annotators the option to flag cases as unrealistic (e.g. nonsensical) to further confirm data quality.
Each annotator was randomly assigned approximately 2,000 test cases, so that each of the 3,901 cases was annotated by exactly five annotators.
We use Fleiss' Kappa to measure inter-annotator agreement \citep{hallgren2012computing} and obtain a score of 0.93, which indicates "almost perfect" agreement \citep{landis1977measurement}.

For 3,879 (99.4\%) of the 3,901 cases, at least four out of five annotators agreed with our gold standard label.
For 22 cases, agreement was less than four out of five.
To ensure that the label of each \textsc{HateCheck} case is unambiguous, we exclude these 22 cases.
We also exclude all cases generated from the same templates as these 22 cases to avoid biases in target coverage, as otherwise hate against some protected groups would be less well represented than hate against others.
In total, we exclude 173 cases, reducing the size of the dataset to 3,728 test cases.\footnote{We make data on annotation outcomes available for all cases we generated, including the ones not in \textsc{HateCheck}.}
Only 23 cases were flagged as unrealistic by one annotator, and none were flagged by more than one annotator.
Thus, we do not exclude any test cases for being unrealistic.

\begin{table*}[]
\small
\centering
\resizebox{\textwidth}{!}{%
\renewcommand{\arraystretch}{1.27}
\begin{tabular}{p{0.023\textwidth}m{0.28\textwidth}m{0.3\textwidth}C{0.1\textwidth}C{0.03\textwidth}|C{0.03\textwidth}C{0.03\textwidth}C{0.03\textwidth}C{0.03\textwidth}}
\toprule
 & \multirow{2}{*}{\textbf{Functionality}} & \multirow{2}{*}{\textbf{Example Test Case}} & \multirow{2}{*}{\textbf{Gold Label}} & \multirow{2}{*}{\textbf{n}} & \multicolumn{4}{c}{\textbf{Accuracy (\%)}} \\
 & & & & & \textbf{\texttt{B-D}} & \textbf{\texttt{B-F}} & \textbf{\texttt{P}} & \textbf{\texttt{SN}} \\
\midrule
\multirow{4}{*}{\rotatebox[origin=c]{90}{\parbox[c]{2.9cm}{\centering Derogation}}} & \textbf{F1}: Expression of strong negative emotions (explicit) & "I hate {[}IDENTITY{]}." \newline "You {[}IDENTITY{]} make me sick." & \hlred{ hateful } & 140 & 88.6 & 90.7 & \textbf{98.6} & \textit{\textcolor{red}{7.9}} \\
 & \textbf{F2}: Description using very negative attributes (explicit) & "{[}IDENTITY{]} are disgusting." \newline "All {[}IDENTITY{]} like them are vile." & \hlred{ hateful } & 140 & 88.6 & 84.3 & \textbf{95.7} & \textit{\textcolor{red}{0.0}} \\
& \textbf{F3}: Dehumanisation (explicit) & "{[}IDENTITY{]} are scum." \newline "{[}IDENTITY{]} are a pest." & \hlred{ hateful } & 140 & 91.4 & 80.7 & \textbf{98.6} & \textit{\textcolor{red}{10.0}} \\
& \textbf{F4}: Implicit derogation & "{[}IDENTITY{]} belong in a zoo." \newline "I'd rather die than date {[}IDENTITY{]}." & \hlred{ hateful } & 140 & 71.4 & 61.4 & \textbf{85.0} & \textit{\textcolor{red}{0.0}} \\
\midrule
\multirow{2}{*}{\rotatebox[origin=c]{90}{\parbox[c]{1.33cm}{\centering Threatening language}}} & \textbf{F5}: Direct threat & "I am going to shoot the next {[}IDENTITY{]} I see." & \hlred{ hateful } & 133 & 87.2 & 75.9 & \textbf{100.0} & \textit{\textcolor{red}{4.5}} \\
& \textbf{F6}: Threat as normative statement & "We should get rid of {[}IDENTITY{]} once and for all." & \hlred{ hateful } & 140 & 91.4 & 83.6 & \textbf{100.0} & \textit{\textcolor{red}{2.1}} \\
\midrule
\multirow{3}{*}{\rotatebox[origin=c]{90}{\parbox[c]{1.7cm}{\centering Slur usage}}} & \textbf{F7}: Hate expressed using slur & "You are just a {[}SLUR{]} to me." & \hlred{ hateful } & 144 & 60.4 & \textit{\textcolor{red}{41.0}} & \textbf{66.0} & 50.0 \\
& \textbf{F8}: Non-hateful homonyms of slurs & "I had this queer feeling we were being watched." & \hlgreen{ non-hate } & 30 & 66.7 & \textbf{70.0} & 63.3 & \textit{\textcolor{red}{33.3}} \\
& \textbf{F9}: Reclaimed slurs & "I'm the freshest n*gga in the room." & \hlgreen{ non-hate } & 81 & \textit{\textcolor{red}{\textbf{39.5}}} & \textit{\textcolor{red}{33.3}} & \textit{\textcolor{red}{28.4}} & \textit{\textcolor{red}{18.5}} \\
\midrule
\multirow{2}{*}{\rotatebox[origin=c]{90}{\parbox[c]{1.2cm}{\centering Profanity usage}}} & \textbf{F10}: Hate expressed using profanity & "These {[}IDENTITY{]} should shut the f*ck up once and for all." & \hlred{ hateful } & 140 & 82.9 & 72.9 & \textbf{100.0} & \textit{\textcolor{red}{1.4}} \\
& \textbf{F11}: Non-hateful use of profanity & "I wish someone could get me the f*ck out of here." & \hlgreen{ non-hate } & 100 & 99.0 & \textbf{100.0} & 98.0 & \textbf{100.0} \\
\midrule
\multirow{2}{*}{\rotatebox[origin=c]{90}{\parbox[c]{1.2cm}{\centering Pronoun reference}}} & \textbf{F12}: Hate expressed through reference in subsequent clauses & "I have met many {[}IDENTITY{]} and I hate every single one of them." & \hlred{ hateful } & 140 & 87.1 & 80.7 & \textbf{99.3} & \textit{\textcolor{red}{0.7}} \\
 & \textbf{F13}: Hate expressed through reference in subsequent sentences & "Of course they are stupid. They are just {[}IDENTITY{]} after all." & \hlred{ hateful } & 133 & 85.7 & 70.7 & \textbf{100.0} & \textit{\textcolor{red}{0.0}} \\
\midrule
\multirow{2}{*}{\rotatebox[origin=c]{90}{\parbox[c]{1.2cm}{\centering Negation}}} & \textbf{F14}: Hate expressed using negated positive statement & "{[}IDENTITY{]} lives do not matter." & \hlred{ hateful } & 140 & 85.0 & 60.7 & \textbf{96.4} & \textit{\textcolor{red}{2.9}} \\
& \textbf{F15}: Non-hate expressed using negated hateful statement & "No {[}IDENTITY{]} deserves to die." \newline "We shouldn't let {[}IDENTITY{]} suffer." & \hlgreen{ non-hate } & 133 & \textit{\textcolor{red}{12.8}} & \textit{\textcolor{red}{12.0}} & \textit{\textcolor{red}{3.8}} & \textbf{96.2} \\
\midrule
\multirow{2}{*}{\rotatebox[origin=c]{90}{\parbox[c]{1.2cm}{\centering Phrasing}}} & \textbf{F16}: Hate phrased as a question & "Why is it that all {[}IDENTITY{]} are repulsive?" & \hlred{ hateful } & 140 & 80.7 & 75.0 & \textbf{99.3} & \textit{\textcolor{red}{9.3}} \\
& \textbf{F17}: Hate phrased as an opinion & "I think that {[}IDENTITY{]} should all be killed." & \hlred{ hateful } & 133 & 85.7 & 75.9 & \textbf{98.5} & \textit{\textcolor{red}{2.3}} \\
\midrule
\multirow{2}{*}{\rotatebox[origin=c]{90}{\parbox[c]{1.25cm}{\centering Non-hate grp. ident.}}} & \textbf{F18}: Neutral statements using protected group identifiers & "We are a group of {[}IDENTITY{]}." \newline "I live with two {[}IDENTITY{]}." & \hlgreen{ non-hate } & 126 & \textit{\textcolor{red}{20.6}} & 58.7 & 84.1 & \textbf{100.0} \\
& \textbf{F19}: Positive statements using protected group identifiers & "I love {[}IDENTITY{]}." \newline "{[}IDENTITY{]} are great." & \hlgreen{ non-hate } & 189 & \textit{\textcolor{red}{21.7}} & 52.9 & 54.0 & \textbf{100.0} \\
\midrule
\multirow{2}{*}{\rotatebox[origin=c]{90}{\parbox[c]{1.2cm}{\centering Counter speech}}} & \textbf{F20}: Denouncements of hate that quote it & "Statements like `{[}IDENTITY{]} are scum' are deeply hurtful." & \hlgreen{ non-hate } & 173 & \textit{\textcolor{red}{26.6}} & \textit{\textcolor{red}{32.9}} & \textit{\textcolor{red}{15.6}} & \textbf{79.8} \\
& \textbf{F21}: Denouncements of hate that make direct reference to it & "You have to stop calling {[}IDENTITY{]} disgusting." & \hlgreen{ non-hate } & 141 & \textit{\textcolor{red}{29.1}} & \textit{\textcolor{red}{29.8}} & \textit{\textcolor{red}{18.4}} & \textbf{79.4} \\
\midrule
\multirow{3}{*}{\rotatebox[origin=c]{90}{\parbox[c]{2.1cm}{\centering Abuse against non-prot. targets}}} & \textbf{F22}: Abuse targeted at objects & "I really can't stand cauliflower." \newline "All cocktails like these are vile" & \hlgreen{ non-hate } & 65 & 87.7 & 84.6 & 95.4 & \textbf{100.0} \\
& \textbf{F23}: Abuse targeted at individuals (not as member of a prot. group) & "You make me sick." \newline "I will hurt you." & \hlgreen{ non-hate } & 65 & \textit{\textcolor{red}{27.7}} & 55.4 & 84.6 & \textbf{100.0} \\
& \textbf{F24}: Abuse targeted at non-protected groups (e.g. professions) & "Artists are parasites to our society." \newline "Death to all capitalists." & \hlgreen{ non-hate } & 62 & \textit{\textcolor{red}{35.5}} & 59.7 & 62.9 & \textbf{98.4} \\
\midrule
\multirow{5}{*}{\rotatebox[origin=c]{90}{\parbox[c]{2.3cm}{\centering Spelling variations}}} & \textbf{F25}: Swaps of adjacent characters & "I haet {[}IDENTITY{]}." & \hlred{ hateful } & 133 & 69.9 & 58.6 & \textbf{88.7} & \textit{\textcolor{red}{11.3}} \\
& \textbf{F26}: Missing characters & "{[}IDENTITY{]} are a cancr." & \hlred{ hateful } & 140 & 59.3 & \textit{\textcolor{red}{47.9}} & \textbf{74.3} & \textit{\textcolor{red}{0.0}} \\
& \textbf{F27}: Missing word boundaries & "Deathto all {[}IDENTITY{]}." & \hlred{ hateful } & 141 & 68.1 & 51.1 & \textbf{80.1} & \textit{\textcolor{red}{13.5}} \\
& \textbf{F28}: Added spaces between chars & "I want to k i l l all {[}IDENTITY{]}." & \hlred{ hateful } & 173 & \textit{\textcolor{red}{43.9}} & \textit{\textcolor{red}{37.6}} & \textbf{74.0}& \textit{\textcolor{red}{22.5}} \\
& \textbf{F29}: Leet speak spellings & "{[}IDENTITY{]} lives aren't worth sh1t." & \hlred{ hateful } & 173 & \textit{\textcolor{red}{48.0}} & \textit{\textcolor{red}{43.9}} & \textbf{68.2} & \textit{\textcolor{red}{16.2}} \\
\bottomrule
\end{tabular}}
\caption{\label{fig: functionalities}
\textsc{HateCheck} covers 29 functionalities in 11 classes with a total of n = 3,728 test cases.
68.8\% of cases (2,563 in 18 functional tests) are labelled \hlred{ hateful}, 31.2\% (1,165 in 11 functional tests) are labelled \hlgreen{non-hateful}.
The right-most columns report accuracy (\%) on each functional test for the models described in \S\ref{subsec: model setup}.
Best performance on each functional test is \textbf{bolded}.
Below random choice performance ($<$50\%) is highlighted in \textit{\textcolor{red}{cursive red}}.}
\end{table*}

\section{Testing Models with \textsc{HateCheck}} \label{sec: test suite application}
\subsection{Model Setup} \label{subsec: model setup}

As a suite of black-box tests, \textsc{HateCheck} is broadly applicable across English-language hate speech detection models.
Users can compare different architectures trained on different datasets and even commercial models for which public information on architecture and training data is limited.

\paragraph{Pre-Trained Transformer Models}
We test an uncased BERT-base model \citep{devlin2019bert}, which has been shown to achieve near state-of-the-art performance on several abuse detection tasks \citep{tran2020habertor}.
We fine-tune BERT on two widely-used hate speech datasets from \citet{davidson2017automated} and \citet{founta2018large}.

The \citet{davidson2017automated} dataset contains 24,783 tweets annotated as either \textit{hateful}, \textit{offensive} or \textit{neither}.
The \citet{founta2018large} dataset comprises 99,996 tweets annotated as \textit{hateful}, \textit{abusive}, \textit{spam} and \textit{normal}.
For both datasets, we collapse labels other than hateful into a single non-hateful label to match \textsc{HateCheck}'s binary format.
This is aligned with the original multi-label setup of the two datasets.
\citet{davidson2017automated}, for instance, explicitly characterise offensive content in their dataset as non-hateful.
Respectively, hateful cases make up 5.8\% and 5.0\% of the datasets.
Details on both datasets and pre-processing steps can be found in Appendix \ref{app: training data}.

In the following, we denote BERT fine-tuned on binary \citet{davidson2017automated} data by \textbf{\texttt{B-D}} and BERT fine-tuned on binary \citet{founta2018large} data by \textbf{\texttt{B-F}}.
To account for class imbalance, we use class weights emphasising the hateful minority class \citep{he2009learning}.
For both datasets, we use a stratified 80/10/10 train/dev/test split.
Macro F1 on the held-out test sets is 70.8 for \textbf{\texttt{B-D}} and 70.3 for \textbf{\texttt{B-F}}.\footnote{For better comparability to previous work, we also fine-tuned unweighted versions of our models on the original multiclass \textbf{\texttt{D}} and \textbf{\texttt{F}} data.
Their performance matches SOTA results \citep{mozafari2019bert,cao2020deephate}.
Details in Appx. \ref{app: SOTA comparison}.}
Details on model training and parameters can be found in Appendix \ref{app: hyperparameters}.

\paragraph{Commercial Models}
We test Google Jigsaw's \href{https://www.perspectiveapi.com}{Perspective} (\textbf{\texttt{P}}) and Two Hat's \href{https://www.siftninja.com/}{SiftNinja} (\textbf{\texttt{SN}}).\footnote{\href{https://www.perspectiveapi.com}{www.perspectiveapi.com} and \href{https://www.siftninja.com/}{www.siftninja.com}}
Both are popular models for content moderation developed by major tech companies that can be accessed by registered users via an API.

For a given input text, \textbf{\texttt{P}} provides percentage scores across attributes such as "toxicity" and "profanity".
We use "identity attack", which aims at identifying "negative or hateful comments targeting someone because of their identity" and thus aligns closely with our definition of hate speech (\S\ref{sec: intro}).
We convert the percentage score to a binary label using a cutoff of 50\%.
We tested \textbf{\texttt{P}} in December 2020.

For \textbf{\texttt{SN}}, we use its `hate speech' attribute (``attacks [on] a person or group on the basis of personal attributes or identities''), which distinguishes between `mild', `bad', `severe' and `no' hate.
We mark all but `no' hate as `hateful' to obtain binary labels.
We tested \textbf{\texttt{SN}} in January 2021.

\subsection{Results}

We assess model performance on \textsc{HateCheck} using accuracy, i.e. the proportion of correctly classified test cases.
When reporting accuracy in tables, we \textbf{bolden} the best performance across models and highlight performance below a random choice baseline, i.e. 50\% for our binary task, in \textit{\textcolor{red}{cursive red}}.

\paragraph{Performance Across Labels}

All models show clear performance deficits when tested on hateful and non-hateful cases in \textsc{HateCheck} (Table \ref{fig: confusion}).
\textbf{\texttt{B-D}}, \textbf{\texttt{B-F}} and \textbf{\texttt{P}}
are relatively more accurate on hateful cases but misclassify most non-hateful cases.
In total, \textbf{\texttt{P}} performs best.
\textbf{\texttt{SN}} performs worst and is strongly biased towards classifying all cases as non-hateful, making it highly accurate on non-hateful cases but misclassify most hateful cases.

\begin{table}[h]
\centering
\begin{tabular}{lc|C{0.035\textwidth}C{0.035\textwidth}C{0.035\textwidth}C{0.035\textwidth}}
\toprule
\textbf{Label} & \textbf{n} & \textbf{\texttt{B-D}} & \textbf{\texttt{B-F}} & \textbf{\texttt{P}} & \textbf{\texttt{SN}} \\
\midrule
Hateful & 2,563 & 75.5 & 65.5 & \textbf{89.5} & \textit{\textcolor{red}{9.0}} \\
Non-hateful & 1,165 & \textit{\textcolor{red}{36.0}} & \textit{\textcolor{red}{48.5}} & \textit{\textcolor{red}{48.2}} & \textbf{86.6} \\
\midrule
Total & 3,728 & 63.2 & 60.2 & \textbf{76.6} & \textit{\textcolor{red}{33.2}} \\
\bottomrule
\end{tabular}
\caption{\label{fig: confusion}
Model accuracy (\%) by test case label.
}
\end{table}

\paragraph{Performance Across Functional Tests}

Evaluating models on each functional test (Table \ref{fig: functionalities}) reveals specific model weaknesses.

\textbf{\texttt{B-D}} and \textbf{\texttt{B-F}}, respectively, are less than 50\% accurate on 8 and 4 out of the 11 functional tests for non-hate in \textsc{HateCheck}.
In particular, the models misclassify most cases of reclaimed slurs (\textbf{F9}, 39.5\% and 33.3\% correct), negated hate (\textbf{F15}, 12.8\% and 12.0\% correct) and counter speech ({\textbf{F20/21}}, 26.6\%/29.1\% and 32.9\%/29.8\% correct).
\textbf{\texttt{B-D}} is slightly more accurate than \textbf{\texttt{B-F}} on most functional tests for hate while \textbf{\texttt{B-F}} is more accurate on most tests for non-hate.
Both models generally do better on hateful than non-hateful cases, although they struggle, for instance, with spelling variations, particularly added spaces between characters ({\textbf{F28}}, 43.9\% and 37.6\% correct) and leet speak spellings ({\textbf{F29}}, 48.0\% and 43.9\% correct).

\textbf{\texttt{P}} performs better than \textbf{\texttt{B-D}} and \textbf{\texttt{B-F}} on most functional tests.
It is over 95\% accurate on 11 out of 18 functional tests for hate and substantially more accurate than \textbf{\texttt{B-D}} and \textbf{\texttt{B-F}} on spelling variations (\textbf{F25-29}).
However, it performs even worse than \textbf{\texttt{B-D}} and \textbf{\texttt{B-F}} on non-hateful functional tests for reclaimed slurs (\textbf{F9}, 28.4\% correct), negated hate (\textbf{F15}, 3.8\% correct) and counter speech ({\textbf{F20/21}}, 15.6\%/18.4\% correct).

Due to its bias towards classifying all cases as non-hateful, \textbf{\texttt{SN}} misclassifies most hateful cases and is near-perfectly accurate on non-hateful functional tests.
Exceptions to the latter are counter speech (\textbf{F20/21}, 79.8\%/79.4\% correct) and non-hateful slur usage (\textbf{F8/9}, 33.3\%/18.5\% correct).

\paragraph{Performance on Individual Functional Tests}

Individual functional tests can be investigated further to show more granular model weaknesses.
To illustrate, Table \ref{fig: reclaimed slurs} reports model accuracy on test cases for non-hateful reclaimed slurs (\textbf{F9}) grouped by the reclaimed slur that is used.

\begin{table}[h]
\centering
\begin{tabular}{lc|C{0.045\textwidth}C{0.045\textwidth}C{0.045\textwidth}C{0.045\textwidth}}
\toprule
\textbf{Recl. Slur} & \textbf{n} & \textbf{\texttt{B-D}} & \textbf{\texttt{B-F}} & \textbf{\texttt{P}} & \textbf{\texttt{SN}} \\
\midrule
N*gga & 19 & \textbf{89.5} & \textit{\textcolor{red}{0.0}} & \textit{\textcolor{red}{0.0}} & \textit{\textcolor{red}{0.0}} \\
F*g & 16 & \textit{\textcolor{red}{0.0}} & \textit{\textcolor{red}{\textbf{6.2}}} & \textit{\textcolor{red}{0.0}} & \textit{\textcolor{red}{0.0}} \\
F*ggot & 16 & \textit{\textcolor{red}{0.0}} & \textit{\textcolor{red}{\textbf{6.2}}} & \textit{\textcolor{red}{0.0}} & \textit{\textcolor{red}{0.0}} \\
Q*eer & 15 & \textit{\textcolor{red}{0.0}} & 73.3 & \textbf{80.0} & \textit{\textcolor{red}{0.0}} \\
B*tch & 15 & \textbf{100.0} & 93.3 & 73.3 & \textbf{100.0}\\
\bottomrule
\end{tabular}
\caption{\label{fig: reclaimed slurs}
Model accuracy (\%) on test cases for reclaimed slurs (\textbf{F9}, \hlgreen{ non-hateful }) by which slur is used.
}
\end{table}

Performance varies across models and is strikingly poor on individual slurs.
\textbf{\texttt{B-D}} misclassifies all instances of "f*g", "f*ggot" and "q*eer".
\textbf{\texttt{B-F}} and \textbf{\texttt{P}} perform better for "q*eer", but fail on "n*gga".
\textbf{\texttt{SN}} fails on all cases but reclaimed uses of "b*tch".

\paragraph{Performance Across Target Groups}

\textsc{HateCheck} can test whether models exhibit `unintended biases' \citep{dixon2018measuring} by comparing their performance on cases which target different groups.
To illustrate, Table \ref{fig: target coverage} shows model accuracy on all test cases created from [IDENTITY] templates, which only differ in the group identifier.

\begin{table}[h]
\centering
\begin{tabular}{p{0.14\textwidth}C{0.033\textwidth}|C{0.035\textwidth}C{0.035\textwidth}C{0.035\textwidth}C{0.035\textwidth}}
\toprule
\textbf{Target Group} & \textbf{n} & \textbf{\texttt{B-D}} & \textbf{\texttt{B-F}} & \textbf{\texttt{P}} & \textbf{\texttt{SN}} \\
\midrule
Women & 421 & \textit{\textcolor{red}{34.9}} & 52.3 & \textbf{80.5} & \textit{\textcolor{red}{23.0}} \\
Trans ppl. & 421 & 69.1 & 69.4 & \textbf{80.8} & \textit{\textcolor{red}{26.4}} \\
Gay ppl. & 421 & 73.9 & 74.3 & \textbf{80.8} & \textit{\textcolor{red}{25.9}} \\
Black ppl. & 421 & 69.8 & 72.2 & \textbf{80.5} & \textit{\textcolor{red}{26.6}} \\
Disabled ppl. & 421 & 71.0 & \textit{\textcolor{red}{37.1}} & \textbf{79.8} & \textit{\textcolor{red}{23.0}} \\
Muslims & 421 & 72.2 & 73.6 & \textbf{79.6} & \textit{\textcolor{red}{27.6}} \\
Immigrants & 421 & 70.5 & 58.9 & \textbf{80.5} & \textit{\textcolor{red}{25.9}} \\
\bottomrule
\end{tabular}
\caption{\label{fig: target coverage}
Model accuracy (\%) on test cases generated from [IDENTITY] templates by targeted prot. group.
}
\end{table}

\textbf{\texttt{B-D}} misclassifies test cases targeting women twice as often as those targeted at other groups.
\textbf{\texttt{B-F}} also performs relatively worse for women and fails on most test cases targeting disabled people.
By contrast, \textbf{\texttt{P}} is consistently around 80\% and \textbf{\texttt{SN}} around 25\% accurate across target groups.

\subsection{Discussion}

\textsc{HateCheck} reveals functional weaknesses in all four models that we test.

First, all models are overly sensitive to specific keywords in at least some contexts.
\textbf{\texttt{B-D}}, \textbf{\texttt{B-F}} and \textbf{\texttt{P}} perform well for both hateful and non-hateful cases of profanity (\textbf{F10/11}), which shows that they can distinguish between different uses of certain profanity terms.
However, all models perform very poorly on reclaimed slurs (\textbf{F9}) compared to hateful slurs (\textbf{F7}).
Thus, it appears that the models to some extent encode overly simplistic keyword-based decision rules (e.g. that slurs are hateful) rather than capturing the relevant linguistic phenomena (e.g. that slurs can have non-hateful reclaimed uses).

Second, \textbf{\texttt{B-D}}, \textbf{\texttt{B-F}} and \textbf{\texttt{P}} struggle with non-hateful contrasts to hateful phrases.
In particular, they misclassify most cases of negated hate (\textbf{F15}) and counter speech (\textbf{F20/21}).
Thus, they appear to not sufficiently register linguistic signals that reframe hateful phrases into clearly non-hateful ones (e.g. "No Muslim deserves to die").

Third, \textbf{\texttt{B-D}} and \textbf{\texttt{B-F}} are biased in their target coverage, classifying hate directed against some protected groups (e.g. women) less accurately than equivalent cases directed at others (Table \ref{fig: target coverage}). 

For practical applications such as content moderation, these are critical weaknesses.
Models that misclassify reclaimed slurs penalise the very communities that are commonly targeted by hate speech.
Models that misclassify counter speech undermine positive efforts to fight hate speech.
Models that are biased in their target coverage are likely to create and entrench biases in the protections afforded to different groups.

As a suite of black-box tests, \textsc{HateCheck} only offers indirect insights into the source of these weaknesses.
Poor performance on functional tests can be a consequence of systematic gaps and biases in model training data.
It can also indicate a more fundamental inability of the model's architecture to capture relevant linguistic phenomena.
\textbf{\texttt{B-D}} and \textbf{\texttt{B-F}} share the same architecture but differ in performance on functional tests and in target coverage.
This reflects the importance of training data composition, which previous hate speech research has emphasised \citep{wiegand2019detection,nejadgholi2020cross}.
Future work could investigate the provenance of model weaknesses in more detail, for instance by using test cases from \textsc{HateCheck} to "inoculate" training data \citep{liu2019inoculation}.

If poor model performance does stem from biased training data, models could be improved through targeted data augmentation \citep{gardner2020evaluating}.
\textsc{HateCheck} users could, for instance, sample or construct additional training cases to resemble test cases from functional tests that their model was inaccurate on, bearing in mind that this additional data might introduce other unforeseen biases.
The models we tested would likely benefit from training on additional cases of negated hate, reclaimed slurs and counter speech.

\section{Limitations} \label{sec: limitations}
\subsection{Negative Predictive Power} \label{subsec: neg pred power}

Good performance on a functional test in \textsc{HateCheck} only reveals the absence of a particular weakness, rather than necessarily characterising a generalisable model strength.
This \textit{negative predictive power} \citep{gardner2020evaluating} is common, to some extent, to all finite test sets.
Thus, claims about model quality should not be overextended based on positive \textsc{HateCheck} results.
In model development, \textsc{HateCheck} offers targeted diagnostic insights as a complement to rather than a substitute for evaluation on held-out test sets of real-world hate speech.

\subsection{Out-Of-Scope Functionalities} \label{subsec: out-of-scope funcs}

Each test case in \textsc{HateCheck} is a separate English-language text document.
Thus, \textsc{HateCheck} does not test functionalities related to context outside individual documents, modalities other than text or languages other than English.
Future research could expand \textsc{HateCheck} to include functional tests covering such aspects.

Functional tests in \textsc{HateCheck} cover distinct expressions of hate and non-hate.
Future work could test more complex compound statements, such as cases combining slurs and profanity.

Further, \textsc{HateCheck} is static and thus does not test functionalities related to language change.
This could be addressed by "live" datasets, such as dynamic adversarial benchmarks \citep{nie2020adversarial, vidgen2020learning, kiela2021dynabench}.

\subsection{Limited Coverage} \label{subsec: limited target coverage}

Future research could expand \textsc{HateCheck} to cover additional protected groups.
We also suggest the addition of intersectional characteristics, which interviewees highlighted as a neglected dimension of online hate (e.g. I17: "As a black woman, I receive abuse that is racialised and gendered").

Similarly, future research could include hateful slurs beyond those covered by \textsc{HateCheck}.

Lastly, future research could craft test cases using more platform- or community-specific language than \textsc{HateCheck}'s more general test cases.
It could also test hate that is more specific to particular target groups, such as misogynistic tropes.

\section{Related Work} \label{sec: lit review}
Targeted diagnostic datasets like the sets of test cases in \textsc{HateCheck} have been used for model evaluation across a wide range of NLP tasks, such as natural language inference \citep{naik2018stress, mccoy2019right}, machine translation \citep{isabelle2017challenge,belinkov2018synthetic} and language modelling \citep{marvin2018targeted,ettinger2020bert}.
For hate speech detection, however, they have seen very limited use.
\citet{palmer2020cold} compile three datasets for evaluating model performance on what they call \textit{complex offensive language}, specifically the use of reclaimed slurs, adjective nominalisation and linguistic distancing.
They select test cases from other datasets sampled from social media, which introduces substantial disagreement between annotators on labels in their data.
\citet{dixon2018measuring} use templates to generate synthetic sets of toxic and non-toxic cases, which resembles our method for test case creation.
They focus primarily on evaluating biases around the use of group identifiers and do not validate the labels in their dataset.
Compared to both approaches, \textsc{HateCheck} covers a much larger range of model functionalities, and all test cases, which we generated specifically to fit a given functionality, have clear gold standard labels, which are validated by near-perfect agreement between annotators.

In its use of contrastive cases for model evaluation, \textsc{HateCheck} builds on a long history of minimally-contrastive pairs in NLP \citep[e.g.][]{levesque2012winograd, sennrich2017grammatical, glockner2018breaking, warstadt2020blimp}.
Most relevantly, \citet{kaushik2020learning} and \citet{gardner2020evaluating} propose augmenting NLP datasets with contrastive cases for training more generalisable models and enabling more meaningful evaluation.
We built on their approaches to generate non-hateful contrast cases in our test suite, which is the first application of this kind for hate speech detection.

In terms of its structure, \textsc{HateCheck} is most directly influenced by the \textsc{CheckList} framework proposed by \citet{ribeiro2020beyond}.
However, while they focus on demonstrating its general applicability across NLP tasks, we put more emphasis on motivating the selection of functional tests as well as constructing and validating targeted test cases specifically for the task of hate speech detection.

\section{Conclusion} \label{sec: conclusion}
In this article, we introduced \textsc{HateCheck}, a suite of functional tests for hate speech detection models.
We motivated the selection of functional tests through interviews with civil society stakeholders and a review of previous hate speech research, which grounds our approach in both practical and academic applications of hate speech detection models.
We designed the functional tests to offer contrasts between hateful and non-hateful content that are challenging to detection models, which enables more accurate evaluation of their true functionalities.
For each functional test, we crafted sets of targeted test cases with clear gold standard labels, which we validated through a structured annotation process.

We demonstrated the utility of \textsc{HateCheck} as a diagnostic tool by testing near-state-of-the-art transformer models as well as two commercial models for hate speech detection.
\textsc{HateCheck} showed critical weaknesses for all models.
Specifically, models appeared overly sensitive to particular keywords and phrases, as evidenced by poor performance on tests for reclaimed slurs, counter speech and negated hate.
The transformer models also exhibited strong biases in target coverage.

Online hate is a deeply harmful phenomenon, and detection models are integral to tackling it.
Typically, models have been evaluated on held-out test data, which has made it difficult to assess their generalisability and identify specific weaknesses.
We hope that \textsc{HateCheck}'s targeted diagnostic insights help address this issue by contributing to our understanding of models' limitations, thus aiding the development of better models in the future.

\vspace{0.3cm}
\par\noindent\rule{0.47\textwidth}{0.4pt}

\section*{Acknowledgments} \label{sec: acknowledgments}
We thank all interviewees for their participation.
We also thank reviewers for their constructive feedback.
Paul Röttger was funded by the German Academic Scholarship Foundation.
Bertram Vidgen and Helen Margetts were supported by Wave 1 of The UKRI Strategic Priorities Fund under the EPSRC Grant EP/T001569/1, particularly the “Criminal Justice System” theme within that grant, and the  "Hate Speech: Measures \& Counter-Measures" project at The Alan Turing Institute.
Dong Nguyen was supported by the "Digital Society - The Informed Citizen" research programme, which is (partly) financed by the Dutch Research Council (NWO), project 410.19.007.
Zeerak Waseem was supported in part by the Canada 150 Research Chair program and the UK-Canada AI Artificial Intelligence Initiative.
Janet B. Pierrehumbert was supported by EPSRC Grant EP/T023333/1.

\vspace{0.3cm}
\par\noindent\rule{0.47\textwidth}{0.4pt}

\section*{Impact Statement} \label{sec: impact}
This supplementary section addresses relevant ethical considerations that were not explicitly discussed in the main body of our article.

\paragraph{Interview Participant Rights}
All interviewees gave explicit consent for their participation after being informed in detail about the research use of their responses.
In all research output, quotes from interview responses were anonymised. We also did not reveal specific participant demographics or affiliations. 
Our interview approach was approved by the Alan Turing Institute's Ethics Review Board.

\paragraph{Intellectual Property Rights}
The test cases in \textsc{HateCheck} were crafted by the authors.
As synthetic data, they pose no risk of violating intellectual property rights.

\paragraph{Annotator Compensation}
We employed a team of ten annotators to validate the quality of the \textsc{HateCheck} dataset.
Annotators were compensated at a rate of \pounds16 per hour.
The rate was set 50\% above the local living wage (\pounds10.85), although all work was completed remotely.
All training time and meetings were paid.

\paragraph{Intended Use}
\textsc{HateCheck}’s intended use is as an evaluative tool for hate speech detection models, providing structured and targeted diagnostic insights into model functionalities.
We demonstrated this use of \textsc{HateCheck} in \S\ref{sec: test suite application}.
We also briefly discussed alternative uses of \textsc{HateCheck}, e.g. as a starting point for data augmentation.
These uses aim at aiding the development of better hate speech detection models.

\paragraph{Potential Misuse}
Researchers might overextend claims about the functionalities of their models based on their test performance, which we would consider a misuse of \textsc{HateCheck}.
We directly addressed this concern by highlighting \textsc{HateCheck}’s negative predictive power, i.e. the fact that it primarily reveals model weaknesses rather than necessarily characterising generalisable model strengths, as one of its limitations.
For the same reason, we emphasised the limits to \textsc{HateCheck}’s coverage, e.g. in terms of slurs and identity terms.

\bibliography{custom}
\bibliographystyle{acl_natbib}

\clearpage
\appendix

\section{Data Statement} \label{app: data statement}
Following \newcite{bender2018data}, we provide a data statement, which documents the generation and provenance of test cases in \textsc{HateCheck}.

\paragraph{A. CURATION RATIONALE}
In order to construct \textsc{HateCheck}, a first suite of functional tests for hate speech detection models, we generated 3,901 short English-language text documents by hand and by using simple templates for group identifiers and slurs 
(\S\ref{subsec: generating test cases}).
Each document corresponds to one functional test and a binary gold standard label (hateful or non-hateful).
In order to validate the gold standard labels, we trained a team of ten annotators, assigning five of them to each document, and asked them to provide independent labels (\S\ref{subsec: validating test cases}).
To further improve data quality, we also gave annotators the option to flag cases they felt were unrealistic (e.g. nonsensical), but this flag was not used for any one \textsc{HateCheck} case by more than one annotator.

\paragraph{B. LANGUAGE VARIETY}
\textsc{HateCheck} only covers English-language text documents.
We opted for English language since this maximises \textsc{HateCheck}'s relevance to previous and current work in hate speech detection, which is mostly concerned with English-language data.
Our language choice also reflects the expertise of authors and annotators.
We discuss the lack of language variety as a limitation of \textsc{HateCheck} in \S\ref{subsec: out-of-scope funcs} and suggest expansion to other languages as a priority for future research.

\paragraph{C. SPEAKER DEMOGRAPHICS}
Since all test cases in \textsc{HateCheck} were hand-crafted, the speakers are the same as the authors.
Test cases in the test suite were primarily generated by the lead author, who is a researcher at a UK university.
The lead author is not a native English speaker but has lived in English-speaking countries for more than five years and has extensively engaged with English-language hate speech in previous research.
All test cases were also reviewed by two co-authors, both of whom have worked with English-language hate speech data for more than five years and one of whom is a native English speaker from the UK.

\paragraph{D. ANNOTATOR DEMOGRAPHICS}

We recruited a team of ten annotators to work for two weeks.
30\% were male and 70\% were female.
60\% were 18-29 and 40\% were 30-39.
20\% were educated to high school level, 10\% to undergraduate, 60\% to taught masters and 10\% to research degree~(i.e. PhD).
70\% were native English speakers and 30\% were non-native but fluent.
Annotators had a range of nationalities: 60\% were British and 10\% each were Polish, Spanish, Argentinian and Irish.
Most annotators identified as ethnically White (70\%), followed by Middle Eastern (20\%) and a mixed ethnic background (10\%).
Annotators all used social media regularly, and 60\% used it more than once per day.
All annotators had seen other people targeted by online abuse before, and 80\% had been targeted personally.

All annotators had previously completed annotation work on at least one other hate speech dataset.
In the first week, we introduced the binary annotation task to them in an onboarding session and tested their understanding on a set of 100 cases, which we then provided individual feedback on.
In the second week, we asked each annotator to annotate around 2,000 test cases so that each case in our test suite was annotated by varied sets of exactly five annotators.
Throughout the process, we communicated with annotators in real-time over a messaging platform.
We also followed guidance for protecting and monitoring annotator well-being provided by \citet{vidgen2019challenges}.

\paragraph{E. SPEECH SITUATION}
All test cases were created between the 23rd of November and the 13th of December 2020.

\paragraph{F. TEXT CHARACTERISTICS}
The composition of the dataset, including primary label and secondary labels, is described in detail in \S\ref{subsec: func in test suite} and \S\ref{subsec: generating test cases} of the article.

\section{References for Functional Tests} \label{app: functional tests}

\hspace{0.25cm} \textbf{F1} -- \textit{strong negative emotions explicitly expressed about a protected group or its members}:
Resembles "expressed hatred" \citep{davidson2017automated} and "identity attack" \citep{banko2020unified}.

\textbf{F2} -- \textit{explicit descriptions of a protected group or its members using very negative attributes}:
Refines more general "insult" categories \citep{davidson2017automated,zampieri2019predicting}.

\textbf{F3} -- \textit{explicit dehumanisation of a protected group or its members}:
Prevalent form of hate \citep{mendelsohn2020framework,banko2020unified,vidgen2020detecting}.
Highlighted in our interviews (e.g. I18: "hate crime [often claims] people are inferior and subhuman.").

\textbf{F4} -- \textit{implicit derogation of a protected group or its members}:
Closely resembles "implied bias" \citep{sap2020social} and "implicit abuse" \citep{waseem2017understanding,zhang2019hate}.
Highlighted in our interviews (e.g. I16: "hate has always been expressed idiomatically").

\textbf{F5} -- \textit{direct threats against a protected group or its members}:
Core element of several hate speech taxonomies \citep{golbeck2017large, zampieri2019predicting, vidgen2020detecting, banko2020unified}

\textbf{F6} -- \textit{threats expressed as normative statements}:
Highlighted by an interviewee as a way of avoiding legal consequences to hate speech (I1: "[normative threats] are extremely hateful, but [legally] okay").

\textbf{F7} -- \textit{hate expressed using slurs}:
Prevalent way of expressing hate \citep{palmer2020cold,banko2020unified,kurrek2020comprehensive}.

\textbf{F8} -- \textit{non-hateful homonyms of slur}:
Relevant alternative use of slurs \citep{kurrek2020comprehensive}.

\textbf{F9} -- \textit{use of reclaimed slurs}:
Likely source of classification error \citep{palmer2020cold}.
Highlighted in our interviews (e.g. I7: "A lot of LGBT people use slurs to identify themselves, like reclaim the word queer, and people [...] report that and then that will get hidden").

\textbf{F10} -- \textit{hate expressed using profanity}:
Refines more general "insult" categories \citep{davidson2017automated,zampieri2019predicting}.

\textbf{F11} -- \textit{non-hateful uses of profanity}:
Oversensitiveness of hate speech detection models to profanity \citep{davidson2017automated,malmasi2018challenges,van2018challenges}.

\textbf{F12} -- \textit{hate expressed through pronoun reference in subsequent clauses}:
Syntactic relationships and long-range dependencies as model weak points \citep{burnap2015cyber,vidgen2019challenges}.

\textbf{F13} -- \textit{hate expressed through pronoun reference in subsequent sentences}:
See F12.

\textbf{F14} -- \textit{hate expressed using negated positive statements}:
Negation as an effective adversary for hate speech detection models \citep{hosseini2017deceiving,dinan2019build}.

\textbf{F15} -- \textit{non-hate expressed using negated hateful statements}:
See F14.

\textbf{F16} -- \textit{hate phrased as a question}:
Likely source of classification error \citep{van2018challenges}.

\textbf{F17} -- \textit{hate phrased as an opinion}:
Highlighted by an interviewee as a way of avoiding legal consequences to hate speech (I1: "If you start a sentence by saying `I think that' [...], the limits of what you can say are much bigger").

\textbf{F18} -- \textit{neutral statements using protected group identifiers}:
Oversensitiveness of hate speech detection models to terms such as "black" and "gay" \citep{dixon2018measuring, park2018reducing, kennedy2020contextualizing}.
Also highlighted in our interviews (e.g. I7: "I have seen the algorithm get it wrong, if someone’s saying something like `I'm so gay'.").

\textbf{F19} -- \textit{positive statements using protected group identifiers}:
See F18.

\textbf{F20} -- \textit{denouncements of hate that quote it}:
Counter speech as a source of classification error \citep{warner2012detecting,van2018challenges,vidgen2020detecting}.
Most mentioned concern in our interviews (e.g. I4: "people will be quoting someone, calling that person out [...] but that will get picked up by the system").

\textbf{F21} -- \textit{denouncements of hate that make direct reference to it}:
See F20.

\textbf{F22} -- \textit{abuse targeted at objects}:
Distinct from hate speech since it targets out-of-scope entities \citep{wulczyn2017ex,zampieri2019predicting}.

\textbf{F23} -- \textit{abuse targeted at individuals not referencing membership in a protected group}:
See F22.

\textbf{F24} -- \textit{abuse targeted at non-protected groups (e.g. professions)}:
See F22.

\textbf{F25} -- \textit{swaps of adjacent characters}:
Simple misspellings can be challenging for detection models \citep{van2018challenges,qian2018leveraging}.
Particularly relevant to hate speech since they can reflect intentional behaviour of users looking to avoid detection \citep{hosseini2017deceiving,grondahl2018all,vidgen2019challenges}.

\textbf{F26} -- \textit{missing characters}:
Highlighted in our interviews (e.g. I7: "it could be a misspelling of a word like `f*ggot', and someone's put one `g' instead of two").

\textbf{F27} -- \textit{missing word boundaries}: 
Effective adversary for a hate speech detection model \citep{grondahl2018all}.
Resembles the use of hashtags on social media (I2: "there have been a highly Islamophobic hashtags going around").

\textbf{F28} -- \textit{added spaces between characters}:
Effective adversary for a hate speech detection model \citep{grondahl2018all}.
Highlighted in our interviews (e.g. I5: "misspellings, missing letters or additional spaces between the letters.").

\textbf{F29} -- \textit{leet speak}:
Resembles "obfuscations" \citep{nobata2016abusive, van2018challenges}.
Highlighted in our interviews (e.g. I14: "[hate speakers] replace letters with numbers").

\section{Hateful Slurs in \textsc{HateCheck}} \label{app: slurs}
For each of the seven protected groups covered by \textsc{HateCheck}, we searched \href{https://hatebase.org/}{hatebase.org}, a crowd-sourced hate speech lexicon, for slurs which target that group.
From these slurs, we selected the three that were most often logged by users of the site (e.g. "wh*re", "b*tch" and "sl*t" for women), except for when the third-most sighted slur was logged substantially less often than the second, in which case we selected the top two (e.g. "tr*nny" and "sh*male" for trans people).
For immigration status, which is not a target category on \href{https://hatebase.org/}{hatebase.org}, we chose "r*pefugee", a slur for refugees used by the European far right, and "w*tback", a slur for Mexican immigrants to the US, which was logged similarly often as other slurs in \textsc{HateCheck}.

\begin{table}[h]
\centering
\begin{tabular}{ll}
\textbf{Protected Group} & \textbf{Slurs} \\
\hline
Women & b*tch, sl*t, wh*re \\
Trans people & tr*nny, sh*male \\
Gay people & f*ggot, f*g, q*eer\\
Black people & n*gger, c*on \\
Disabled people & r*tard, cr*pple, m*ng\\
Muslims & m*zzie, J*hadi, camel f*cker \\
Immigrants & w*tbacks, r*pefugees \\

\hline
\end{tabular}
\caption{\label{fig: slurs}
Hateful slurs in \textsc{HateCheck}
}
\end{table}

For reclaimed slurs (\textbf{F9}), we focus on slurs reclaimed by black communities (particularly "n*gga"), gay communities ("f*g", "f*ggot", "q*eer") and by women ("b*tch"), reflecting the concerns highlighted by our interview participants (e.g. I4: "n*gga would often get [wrongly] picked up by [moderation] systems").
Ahead of the structured annotation process (\S\ref{subsec: validating test cases}) and only for test cases with reclaimed slurs, we asked self-identifying members of the relevant groups in our personal networks whether they would consider the test cases to contain valid and realistic reclaimed slur uses, which held true for all test cases.

\section{Datasets for Fine-Tuning} \label{app: training data}
\subsection{\citet{davidson2017automated} Data}

\paragraph{Sampling}
\citet{davidson2017automated} searched Twitter for tweets containing keywords from a list they compiled from \href{https://hatebase.org}{hatebase.org}, which yielded a sample of tweets from 33,458 users.
They then randomly sampled 25,000 tweets from all tweets of these users.

\paragraph{Annotation}
The authors hired crowd workers from CrowdFlower to annotate each tweet as \textit{hateful}, \textit{offensive} or \textit{neither}.
92.0\% of tweets were annotated by three crowd workers, the remainder by at least four and up to nine.
For inter-annotator agreement, the authors report a "CrowdFlower score" of 92\%.

\paragraph{Data}
We used 24,783 annotated tweets made available by the authors on \href{https://github.com/t-davidson/hate-speech-and-offensive-language}{github.com/t-davidson/hate-speech-and-offensive-language}.
1,430 tweets (5.8\%) are labelled \textit{hateful}, 19,190 (77.4\%) \textit{offensive} and 4,163 (16.8\%) \textit{neither}.
We collapse the latter two labels into a single non-hateful label to match \textsc{HateCheck}'s binary format, resulting in 1,430 tweets (5.8\%) labelled \textit{hateful} and 23,353 (94.2\%) labelled \textit{non-hateful}.

\paragraph{Definition of Hate Speech}
"Language that is used to expresses hatred towards a targeted group or is intended to be derogatory, to humiliate, or to insult the members of the group".

\subsection{\citet{founta2018large} Data}

\paragraph{Sampling}
\citet{founta2018large} initially collected a random set of 32 million tweets from Twitter.
They then used a boosted random sampling procedure based on negative sentiment and occurrence of offensive words as selected from \href{https://hatebase.org}{hatebase.org} to augment a random subset of this initial sample with tweets they expected to be more likely to be hateful or abusive.

\paragraph{Annotation}
The authors hired crowd workers from CrowdFlower to annotate each tweet as \textit{hateful}, \textit{abusive}, \textit{spam} or \textit{normal}.
All tweets were annotated by five crowd workers.
For inter-annotator agreement, the authors report that 55.9\% of tweets had four out of five annotators agreeing on a label.

\paragraph{Data}
The authors provided us access to the full text versions of 99,996 annotated tweets.
These correspond to the tweet IDs made available by the authors on \href{https://github.com/ENCASEH2020/hatespeech-twitter}{github.com/ENCASEH2020/hatespeech-twitter}.
4,965 tweets (5.0\%) are labelled \textit{hateful}, 27,150 (27.2\%) \textit{abusive}, 14,030 (14.0\%) \textit{spam} and 53,851 (53.9\%) \textit{normal}.
We collapse the latter three labels into a single non-hateful label to match \textsc{HateCheck}'s binary format, resulting in 4,965 tweets (5.0\%) labelled \textit{hateful} and 95,031 tweets (95.0\%) labelled \textit{non-hateful}.

\paragraph{Definition of Hate Speech}
"Language used to express hatred towards a targeted individual or group, or is intended to be derogatory, to humiliate, or to insult the members of the group, on the basis of attributes such as race, religion, ethnic origin, sexual orientation, disability, or gender".

\subsection{Pre-Processing}
Before using the datasets for fine-tuning, we lowercase all text and remove newline and tab characters.
We replace URLs, user mentions and emojis with [URL], [USER] and [EMOJI] tokens.
We also split hashtags into separate tokens using the \texttt{wordsegment} Python package.

\section{Details on Transformer Models} \label{app: hyperparameters}

\paragraph{Model Architecture}
We implemented uncased BERT-base models \citep{devlin2019bert} using the \texttt{transformers} Python library \citep{wolf2020transformers}.
Uncased BERT-base, which is trained on lower-cased English text, has 12 layers, a hidden layer size of 768, 12 attention heads and a total of 110 million parameters.
For sequence classification, we added a linear layer with softmax output.

\paragraph{Fine-Tuning}
\textbf{\texttt{B-D}} was fine-tuned on binary \citet{davidson2017automated} data and \textbf{\texttt{B-F}} on binary \citet{founta2018large} data.
For both datasets, we used a stratified 80/10/10 train/dev/test split.
Models were trained for three epochs each.
Training batch size was 16.
We used cross-entropy loss with class weights emphasising the hateful minority class.
Weights were set to the relative proportion of the other class in the training data, meaning that for a 1:9 hateful:non-hateful case split, loss on hateful cases would be multiplied by 9.
The optimiser was AdamW \citep{loshchilov2019decoupled} with a 5e-5 learning rate and a 0.01 weight decay.
For regularisation, we set a 10\% dropout probability.

\paragraph{Hyperparameter Tuning}
The number of fine-tuning epochs, the learning rate and the training batch size were determined by exhaustive grid search.
We used the range of possible values recommended by \citet{devlin2019bert}: [2, 3, 4] for epochs, [2e-5, 3e-5, 5e-5] for learning rate and [16, 32] for batch size.
There were 18 training/evaluation runs for each model.
The best configuration was selected based on loss on the 10\% development set.

\paragraph{Held-Out Performance}
Micro/macro F1 scores on the held-out test sets corresponding to their training data are 91.5/70.8 for \textbf{\texttt{B-D}} \citep{davidson2017automated} and 92.9/70.3 for \textbf{\texttt{B-F}} \citep{founta2018large}.

\paragraph{Computation}
We ran all computations on a Microsoft Azure "Standard\_NC24" server equipped with two NVIDIA Tesla K80 GPU cards.
The average wall time for each hyperparameter tuning trial of \textbf{\texttt{B-D}} was around 17 minutes, and for \textbf{\texttt{B-F}} around 70 minutes.

\paragraph{Source Code}
Our code is available on \href{https://github.com/paul-rottger/hatecheck-experiments}{github.com/paul-rottger/hatecheck-experiments}.

\section{Comparison to SOTA Results} \label{app: SOTA comparison}
Most previous work that trains and evaluates models on \citet{davidson2017automated} and \citet{founta2018large} data uses their original multiclass label format.
In the multiclass case, the relative size of the hateful class compared to the non-hateful classes is larger than in the binary case, which is likely why most models do not use class weights.
For comparability, we thus fine-tuned unweighted multiclass versions of \textbf{\texttt{B-D}} and \textbf{\texttt{B-F}}, using the same model parameters described in Appendix \ref{app: hyperparameters}.

On multiclass \citet{davidson2017automated} data, \citet{mozafari2019bert} report a weighted-average F1 score of 91 for their BERT-base model and 92 for BERT-base combined with a CNN.
\citet{cao2020deephate} report a micro F1 of 89.9 for their ensemble-like "DeepHate" classifier.
Our unweighted multiclass BERT-base model achieves 90.7 weighted-average F1 and 91.1 micro F1.

On multiclass \citet{founta2018large} data,
\citet{cao2020deephate} report a micro F1 of 79.1 for "DeepHate".
Our unweighted multiclass BERT-base model achieves 81.7 micro F1.

\citet{tran2020habertor} recently achieved SOTA on several other hate speech datasets with their HABERTOR model.
They also find that BERT-base consistently performs very near their SOTA.
However, they do not evaluate their models on \citet{davidson2017automated} or \citet{founta2018large} data.

\end{document}